# A DRIVER ADVISORY SYSTEM BASED ON LARGE LANGUAGE MODEL FOR HIGH-SPEED TRAIN

**Yuchen Luo**
School of Automation and Intelligence
Beijing Jiaotong University, Beijing, China, 100044
Email: 23120238@bjtu.edu.cn

**Jing Xun**
School of Automation and Intelligence
Beijing Jiaotong University, Beijing, China, 100044
Email: jxun@bjtu.edu.cn

**Wei Wang**
China Machinery International Engineering Design & Research Institute Co. Ltd
Changsha, China, 410021
Email: wangwei@cmie.cn

**Ruize Zhang**
Jeme Tienyow Honors College
Beijing Jiaotong University, Beijing, China, 100044
Email: 22331088@bjtu.edu.cn

**Zicong Zhao**
School of Automation and Intelligence
Beijing Jiaotong University, Beijing, China, 100044
Email: 23111076@bjtu.edu.cn

Word Count: 4770 words + 4 table(s) × 250 = 5770 words

Submission Date: January 14, 2025

``

**ABSTRACT**

With the rapid development of China high-speed railway, drivers face increasingly significant technical challenges during operations, such as fault handling. Currently, drivers depend on the onboard mechanic when facing technical issues, for instance, traction loss or sensor faults. This dependency can hinder effective operation, even lead to accidents, while waiting for faults to be addressed. To enhance the accuracy and explainability of actions during fault handling, an Intelligent Driver Advisory System (IDAS) framework based on a large language model (LLM) named IDAS-LLM, is introduced. Initially, domain-fine-tuning of the LLM is performed using a constructed railway knowledge question-and-answer dataset to improve answer accuracy in railway-related questions. Subsequently, integration of the Retrieval-augmented Generation (RAG) architecture is pursued for system design to enhance the explainability of generated responses. Comparative experiments are conducted using the constructed railway driving knowledge assessment dataset. Results indicate that domain-fine-tuned LLMs show an improvement in answer accuracy by an average of 10%, outperforming some current mainstream LLMs. Additionally, the inclusion of the RAG framework increases the average recall rate of question-and-answer sessions by about 4%. Finally, the fault handling capability of IDAS-LLM is demonstrated through simulations of real operational scenarios, proving that the proposed framework has practical application prospects.

# INTRODUCTION

High-speed train drivers play a crucial safety role in the railway transport system, and their training process is both rigorous and complex (*1*). Initially, candidates must accumulate 200,000 kilometers of driving experience on conventional trains over three years to qualify for high-speed train driver training. During the training phase, they are required to study 20 specialized courses, including "High-Speed Trainset Technology", and pass related theoretical exams. Subsequently, they enter a practical training phase at a training base equipped with advanced simulators for high-speed trainsets, such as the CR (China Railway) and CRH (China Railway High-speed) series, which can simulate various emergency scenarios that may occur in real operations. Only when candidates have completed the required mileage and passed practical assessments are they eligible to apply for a high-speed train driving license. The entire training period can span several years. With the rapid development of China's high-speed railways, the demand for train drivers is increasing, and the current system, which often implements a single driver per shift, not only increases the workload (*2*) but may also lead to driver fatigue (*3*), posing a potential threat to operational safety.

Currently, high-speed trains in China primarily operate under manual control. When issues such as traction loss or sensor faults occur, drivers immediately notify the onboard mechanic. The onboard mechanic is responsible for monitoring the operational status of the trainset and handling any emergencies. Once alerted, they promptly address the issue. This arrangement has the advantage of clear roles, but a drawback is that it can place the driver in a dilemma. In a real scene, the train was accelerating to pass through a neutral zone, and the driver got a warning for traction loss. He noticed the mechanic immediately. Then, the situation at that time was, on one hand, the mechanic is busy on solving the fault. On the other hand, the driver needs to cooperate with the mechanic before deciding on if he could go on accelerating the train to the speed required to pass through the neutral zone. However, because the mechanic cannot be reached in a timely manner, the driver missed the time window of acceleration. This caused the accident that the train is unable to pass through the phase zone and stops in it. In such cases, providing drivers with an assistant capable of helping manage fault scenarios could be a solution.

The driver advisory system(DAS) can assist drivers in real-time to provide personalized driving assistance. For instance, it can recommend the best speed based on traffic conditions, weather, and other factors (*4*). In addition, it can help drivers adjust their driving behavior to optimize fuel efficiency, prevent accidents, and comply with traffic regulations. A DAS based on LLM has potential for providing context-specific driving advice to avoid such an accident as aforementioned. To make it trusted by drivers, it needs to address the challenge of improving the explainability of LLM.

In recent years, large language models (LLMs) like ChatGPT (*5*) have demonstrated exceptional capabilities in common sense reasoning, inference, and planning, providing insightful suggestions and finding practical applications in fields such as law (*6*), finance (*7*), medicine (*8*) and road traffic (*9*). While one of potential applications of ChatGPT is intelligent driver assistance (*10*), there are several technical questions that need to be addressed. Most LLMs are pretrained on general corpora and may not cover or accurately represent domain-specific knowledge in railway. Applying LLMs directly to IDAS could result in inaccurate or erroneous outputs. To overcome potential hallucination issues with LLMs in railway knowledge, domain-specific fine-tuning is necessary to enhance performance in fault handling scenarios.

Therefore, the IDAS-LLM framework, a large language model based system designed to assist in managing frequent high-speed train faults, is proposed. Specifically, we begin by con-sulting the Chinese Railway Locomotive and Vehicle Driver's Qualification Examination syllabus and employ existing high-performance large language models to automatically generate a domain-specific fine-tuning dataset for the railway sector. This dataset includes 10,100 structured question-and-answer pairs, covering legal provisions, railway regulations, and railway expertise. Subsequently, we conduct domain-specific fine-tuning using the assembled Railway Training Dataset (RTD) to enhance the LLM's expertise in the railway field. To further address potential hallucination issues within this sector, we integrate Retrieval-Augmented Generation (RAG) technique to improve the accuracy and professionalism of the LLM's responses. We validate the framework on our custom railway driving knowledge assessment dataset and perform comparative evaluations against several contemporary mainstream models. Finally, we demonstrate the framework's response effectiveness through examples of faults encountered in actual high-speed train operations.

The main contributions of this paper are as follows:
- We construct a domain-specific fine-tuning dataset for the railway sector, RTD, consisting of 10,100 structured question-and-answer pairs that cover railway legal provisions, regulations, and expertise.
- We propose a Chinese railway domain-specific driver advisory LLM named IDAS-LLM. The fault handling capabilities of IDAS-LLM are verified under simulated real-world scenarios.
- We introduce RAG technology, which consults a driver knowledge base to mitigate LLM hallucinations, enhancing both the accuracy and explainability of LLM responses.

The rest of this paper is organized as follows. Section 2 reviews current research on IDAS and the application of LLMs in the transportation sector. Section 3 describes the construction process and the technical methods used in IDAS-LLM. Section 4 presents the experimental design and results, including fine-tuning comparison experiments and RAG comparison experiments. Section 5 demonstrateshow IDAS-LLM handles two types of fault scenarios under simulated real-world conditions. Section 6 concludes the paper and proposes future research directions.

**LITERATURE REVIEW**

The current intelligent driver advisory systems are primarily developed with a focus on energy conservation and punctuality, concentrating largely on exploring theories and control methods related to optimizing train operation curves. Zhu et al.(*11*) introduce a novel DAS prototype design that utilizes a PC as the central unit and a smartphone as the on-board unit, enabling bi-directional message exchange between the two units. They thoroughly detail the comprehensive process from the system structure and methodological design to the practical application using a PC with a smartphone, effectively translating research concepts into real-world implementation. Dong et al. (*12*) explore methods to improve train operational performance using DAS, employing enhanced train dynamics and energy efficiency models. They design a traction-distance-based method for trajectory optimization and construct a hardware-in-the-loop experiment platform to simulate the application of DAS in actual driving scenarios. A prototype system is developed to provide essential advisory information to train drivers. Xiao et al. (*13*) propose an onboard energy-efficient driving advisory system for high-speed trains operating on vehicle control units. They introduce a data interaction framework to monitor the train's real-time state and temporary speed limits. Additionally, a bilevel optimal control method is designed to compute the energy-optimal speed profile in realtime, extracting driving advice from the optimized trajectories. Formato et al. (*14*) address the energy aspect with considerations for Timetable programming and TMS regulations.

Specifically, they focus on defining appropriate speed profiles to optimize TMS time constraints and on developing a C-DAS architecture that can be immediately implemented in the current infrastructure. In addition to designs aimed at energy optimization, Wang et al. (*15*) propose a novel algorithm architecture for a connected DAS, designed to provide time/speed advice to freight train drivers. This system offers appropriate recommendations based on a predicted feasible merging window, aiding drivers in achieving smooth merging operations. Guerra et al. (*16*) explore the interaction between train drivers and an autonomous driver advisory system (ADAS) based on human state estimation, aimed at avoiding confrontational interactions and promoting integrative cooperation. The findings indicate that such interactions enhance driver acceptance rates for ADAS recommendations and improve overall train control performance, ensuring mission success.

Large language models are demonstrating their extensive influence across various fields, including road transportation. Research involving these models is gradually unfolding within the road traffic sector, showcasing their potential application value and broad research interest. The current focus of research primarily centers on constructing frameworks driven by large language models. Cui et al.(*17*) present DriveLLM, a decision-making framework that integrates LLMs with existing autonomous driving stacks. This integration enables commonsense reasoning within the decision-making process. DriveLLM also features a unique cyberphysical feedback system, allowing it to learn from its mistakes and make improvements. Zhang et al. (*9*) introduce the TrafficGPT framework, which bridges the critical gap between LLMs and traffic foundation models (TFMs) by defining a series of prompts. This framework aims to infuse LLMs with the capability to interact with traffic data and systems, ensuring reliability, and provides valuable decision support for managing urban transportation systems. Cui et al. (*18*) significantly enhance the decision-making capabilities of autonomous vehicles by deploying LLMs. Their developed framework integrates LLMs' advanced language comprehension and contextual analysis, coupled with strategic tool applications and collaborative operations with various vehicle modules, markedly improving the safety and efficiency of autonomous driving technologies. Additionally, researchers are employing fine-tuning techniques to facilitate knowledge injection into these models. Wang et al. (*19*) introduce TransGPT, a novel (multi-modal) large language model tailored for the transportation domain, comprising two independent variants: TransGPT-SM for single-modal data and TransGPT-MM for multi-modal data. They also showcase the potential applications of TransGPT in traffic analysis and modeling.

**METHODS**

This section explores the construction of the IDAS-LLM, which involves three processes: the construction of the railway driving dataset, supervised fine-tuning (SFT) and the design of RAG. We will discuss each step in sequence to reflect the workflow of the research.

**Construction of Railway Driving Dataset**
*Dataset for Training*

To construct a dataset that accurately reflects the real-world driving scenarios faced by train drivers, we referenced the syllabus of the Chinese Railway Locomotive and Rolling Stock Drivers' Qualification Examination. This ensures that the included knowledge and skills are aligned with industry standards. The dataset is divided into three main categories, each directly corresponding to the core content of the examination syllabus. This categorization aims to provide rich examples and contexts for the subsequent fine-tuning of the large model. The specific categories are as

follows:
- **Legal Provisions:** This category covers laws and regulations related to railway safety and operations, as well as specific ordinances and methods for railway safety management. This content helps train drivers understand their legal responsibilities and ensures compliance with relevant laws and regulations during their work.
- **Railway Regulations:** This category includes specific railway technical operations and maintenance procedures. These regulations are the technical foundation for the effective operation of the railway system, providing standardized operational guidelines to ensure that drivers can correctly perform technical operations and maintenance tasks.
- **Railway Expertise:** This category involves the operation and fault handling techniques of trainsets, as well as emergency handling manuals for various types of trainset failures during transit. This content aims to enhance train drivers' mastery of high-speed trainset technical details and their ability to respond in emergency situations.

The original text data collected comprises a total of 776,000 Chinese tokens, which are used to construct a railway driving text database. Based on this, a railway driving question and answer dataset for fine-tuning training, called the Railway Training Dataset (RTD), will be established. The specific composition of this dataset is shown in Table 1.

**TABLE 1 Statistics of Sources for Railway Fine-Tuning Training Data**

| Category | Total Tokens | Number of Q&A |
|---|---|---|
| Legal Provision | 138,809 | 4,553 |
| Railway Regulation | 371,914 | 3,052 |
| Railway Expertise | 265,634 | 2,517 |

Ensuring that large language models effectively absorb railway domain knowledge and demonstrate conversational abilities is crucial. This process relies on a substantial amount of instructive data, enabling LLM to generate appropriate outputs based on predetermined instructions and inputs. However, there is a significant lack of such datasets in the railway domain, and manually formatting these data is not only time-consuming but also labor-intensive. To address this challenge, an automated processing strategy has been adopted, as shown in Figure 1. Existing large language models with excellent performance are utilized to automatically generate fine-tuning training data. The generated data is primarily stored in a question-and-answer format, facilitating learning and application by the LLM.

Initially, a chunking strategy is used to process the text in the database, segmenting the text into themes based on specific formatting elements such as clauses and chapter titles. Subsequently, leveraging the advanced language understanding capabilities of large language models, suitable prompts are designed and combined with few-shot learning techniques to generate one or more related questions for each text block. For each question, the corresponding text block content is restructured into a prompt input, which is submitted to a large language model to generate answers that accurately reflect key information. During this process, issues such as question duplication, missing, or invalid answers may arise, necessitating strict data filtering before the final integration. After completing the filtering, the question-and-answer pairs are integrated and stored in structured data, thus completing the construction of the RTD and providing data support for conversations in railway driving scenarios.

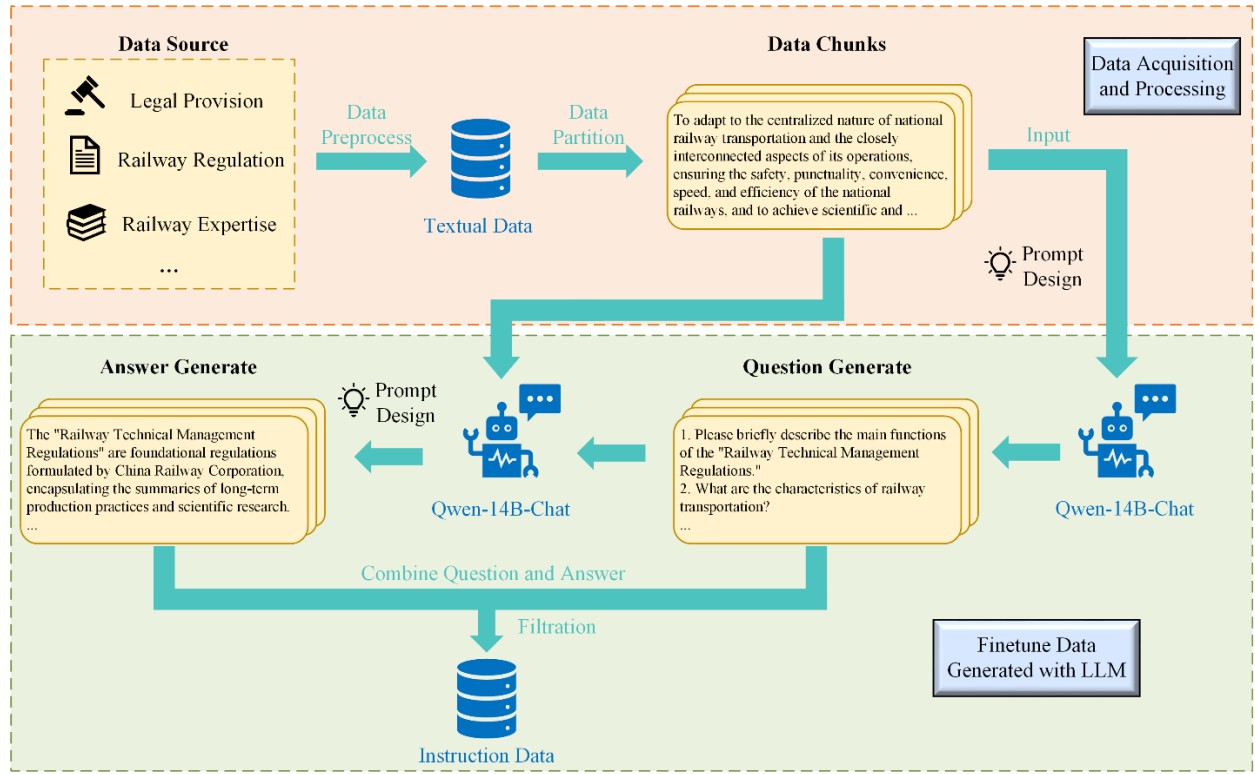

**FIGURE 1 Automated Generation Process for Railway Fine-Tuning Training Data**

In conclusion, using the Qwen-14B-Chat (*20*) model on a single A40 GPU, 10,100 structured question-and-answer pairs are generated for fine-tuning.

*Dataset for Evaluation*

To assess the IDAS-LLM's effectiveness in applying train driver knowledge, an assessment set for railway driving knowledge is created by organizing an existing high-speed train driver assessment question bank. Similarly, by using the Qwen-14B-Chat model, single choice questions, multiple choice questions, and true/false questions from the question bank are transformed into a question-and-answer format to facilitate the evaluation of the framework's text generation performance. The question in each question-and-answer pair is a direct rephrasing of the question from the question bank, while the corresponding reference answer is a rephrased text that combines the question and its corresponding reference answer from the question bank.

A total of 2462 questions are collected, including 845 single choice questions, 642 multiple choice questions, and 975 true/false questions. The data converted into question-and-answer format is categorized by question type, and 100 question-and-answer pairs are extracted from each category for the assessment set. Thus, the railway driving knowledge assessment set includes 100 question-and-answer pairs on legal provisions, 100 on railway regulations, and 100 on professional railway knowledge.

**Supervised Fine-Tuning**

*Model Selection*

SFT is a critical technique for imparting domain-specific knowledge to models, and its effectiveness heavily depends on the selection of the base model. Given that not all models support Chinese, compatibility with the Chinese language becomes a primary consideration in choosing the base model. To evaluate model performance, Chinese large language model evaluation datasets, CEval (*21*) and CMMLU (*22*), will be utilized. These datasets encompass a wide range of academic subjects and are capable of testing the model's comprehension across various domains and its ability to generate natural language. Details on the selection of the base model and its performance scores on the corresponding benchmark datasets are provided in Table 2. In Table 2, the displayed scores represent the optimal values for each model across various evaluation metrics, with the test scores for C-Eval and CMMLU sourced from the official evaluation results or leaderboard submissions for each model.

**TABLE 2  Comparison of Base Model Performance**

| Base Model | C-Eval | CMMLU |
|---|---|---|
| ChatGLM3-6B (*23*) | 69 | 67.5 |
| Baichuan-7B (*24*) | 42.8 | 42.33 |
| Qwen-7B (*20*) | 59.6 | 57.57 |
| Bloomz-7B (*25*) | 35.7 | 42.80 |

Based on the scores, ChatGLM3-6B demonstrates superior performance among base models with less than 10 billion parameters, and thus it has been selected as the base model for IDAS-LLM. In terms of semantic understanding, due to training on a large-scale corpus, this model exhibits strong semantic analysis capabilities. It is able to accurately capture user intent and provide precise responses.

*Training Detail*

Fine-tuning training is conducted on a single A40 GPU. Utilizing the transformers and peft libraries, parameter-efficient fine-tuning is achieved through the Low-rank Adaptation (LoRA) method, with settings following the recommendations found in (*26*). The dataset is divided into training and validation sets at an 8:2 ratio. The model undergoes fine-tuning for 10 epochs, with a learning rate of 5e-4 and a batch size of 4, employing the Adam optimizer and the cross-entropy loss function.

**Design of Retrieval-Augmented Generation**

Given that only 0.03% of the total parameters are trained during fine-tuning, with the trainable parameters in the LoRA algorithm comprising 2M out of the model's total 6B parameters, and considering the influence of dataset quality and quantity, the fine-tuned model is still able to effectively address relevant questions within the railway domain. However, it still exhibits hallucinations regarding certain detailed information and may even produce incorrect responses. Therefore, to further mitigate LLM hallucinations and enhance the accuracy and explainability of the answers, RAG has been introduced. RAG enhances LLMs by retrieving relevant document blocks from an external knowledge base through semantic similarity calculations. By referencing external knowledge, RAG effectively reduces the generation of factually incorrect content. The implementation of RAG is illustrated in Figure 2.

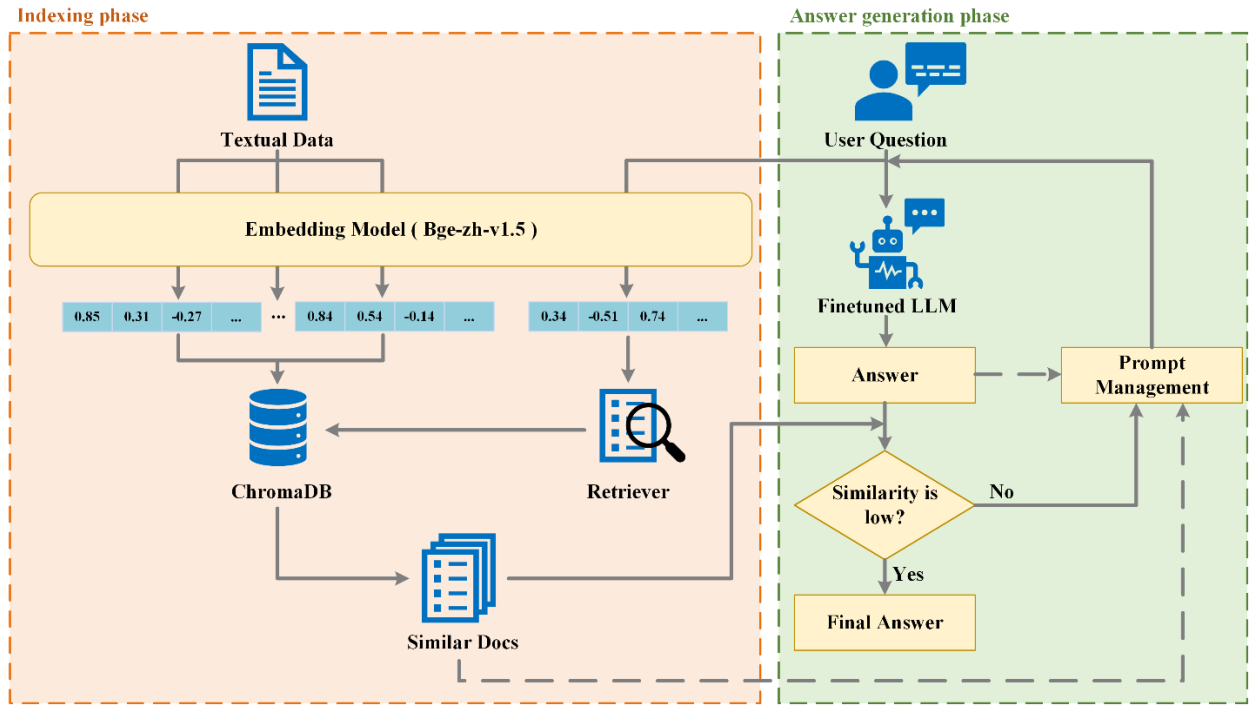

**FIGURE 2 Implementation of RAG in IDAS-LLM. This also constitutes the overall framework of IDAS-LLM.**

Specifically, during the indexing phase, railway textual data are first loaded and processed in discrete chunks. Each chunk is then annotated with its specific source, which is derived from the textual data extracted during the training set generation. Then, utilizing the Bge-zh-v1.5 (*27*) embedding model, these railway text data are vectorized. Once vectorization is complete, the data are stored in the ChromaDB database, thereby constructing a knowledge base of driving data and generating a corresponding retriever. In the answer generation phase, user questions are processed in two ways: one is direct input into the fine-tuned model to generate answers, and the other uses the same Bge-zh-v1.5 embedding model as in the indexing phase to generate a vectorized query. For the latter, the system calculates similarity scores between the generated query and text vectors in the driving data knowledge base, ranking texts by similarity and retrieving the top five texts with the highest similarity to the query. If the similarity score is below a preset threshold, it is assumed that irrelevant texts have been queried, and the answer from the first method of processing is directly output. Conversely, if relevant texts are retrieved, these text blocks are used as extended context in the prompt and re-entered into the fine-tuned model. Combining the answer from the first method, the LLM rethinks and outputs a revised or refined answer, including information about the documents it referenced.

**EXPERIMENTS**

　　　This section conducts experimental validation on the fine-tuning and RAG design within IDAS-LLM, discussing the results of the two experiments through text generation-related evaluation metrics.

**Fine-tuning Comparison Experiments**

　　　Experiments are conducted on the previously prepared railway driving knowledge assessment set. To evaluate the LLM's ability to answer questions across three categories before and after fine-tuning, BLEU (*28*) and ROUGE (*29*) are used as evaluation metrics to measure the differences between model-generated responses and reference responses. BLEU (B) is a precision-based similarity measure that focuses on accuracy, with higher BLEU scores indicating higher accuracy. ROUGE, on the other hand, is a recall-based similarity measure focusing on recall, where higher ROUGE scores indicate higher recall. ROUGE-1 (R1), ROUGE-2 (R2), and ROUGE-L (RL) are chosen as specific metrics for ROUGE, and comparisons are made with some of the current mainstream Chinese large language models. The evaluation results are shown in Table 3.

　　　The results in Table 3 indicate that, compared to the baseline model, the model fine-tuned for the railway domain shows improved performance in answering questions across three categories. Particularly in railway expertise questions, the three ROUGE metrics increased by 8%, 6%, and 3% respectively, and BLEU increased by 7%, indicating that the fine-tuned model has significantly improved at the word level, capable of understanding and responding with professional railway terminology. In legal provisions and railway regulations questions, the fine-tuned model also shows improvements in various metrics. Additionally, the BLEU scores across the three categories have significantly improved after fine-tuning, suggesting that the LLM's responses have become more accurate and capable of capturing relevant information. Compared to current mainstream large Chinese language models, the fine-tuned model performs modestly in handling legal provisions and railway regulations questions, possibly reflecting that these models have already acquired related knowledge during the pre-training phase. Despite this, the fine-tuned model excels in railway professional knowledge, demonstrating that targeted fine-tuning can significantly enhance the domain capability of large language models.

**TABLE 3 Comparison of Evaluation Metrics for Model-generated Responses**

| Models | Legal Provision | | | | Railway Regulation | | | | Railway Expertise | | | |
|---|---|---|---|---|---|---|---|---|---|---|---|---|
| | R1 | R2 | RL | B | R1 | R2 | RL | B | R1 | R2 | RL | B |
| ChatGLM3-6B | 0.44 | 0.19 | 0.37 | 0.06 | 0.36 | 0.14 | 0.31 | 0.06 | 0.30 | 0.10 | 0.25 | 0.03 |
| Internlm2-chat-7b (*30*) | 0.37 | 0.14 | 0.31 | 0.03 | 0.37 | 0.14 | **0.32** | 0.02 | 0.34 | 0.11 | **0.28** | 0.02 |
| Qwen2-7B-Instruct (*31*) | **0.54** | 0.24 | **0.43** | 0.05 | **0.47** | 0.18 | 0.31 | 0.05 | 0.37 | 0.11 | 0.27 | 0.03 |
| Baichuan2-7B-chat (*32*) | 0.42 | 0.17 | 0.34 | 0.06 | 0.37 | 0.14 | **0.32** | 0.05 | 0.30 | 0.10 | 0.25 | 0.04 |
| **Finetuned LLM** | 0.47 | **0.25** | 0.40 | **0.21** | 0.39 | **0.20** | 0.32 | **0.14** | 0.38 | **0.16** | 0.28 | **0.10** |
| Δ% (Baseline) | 3% | 6% | 3% | 15% | 3% | 6% | 1% | 8% | 8% | 6% | 3% | 7% |

　　　Based on the experimental results, ROUGE and BLEU provide effective benchmarks for improvement. However, it is noteworthy that the evaluation metrics for the selected models generally appear low during testing, especially the BLEU scores. This may be due to two factors. On one hand, although these models match the reference answers well in terms of railway or legal terminology, the length of the generated or reference answers leads to lower calculated evaluation metrics. On the other hand, since ROUGE and BLEU primarily focus on literal overlap and do not consider the meanings of words, different words with the same meaning or variants of the same word might be mistakenly judged as incorrect generations.

**RAG comparison Experiments**

To evaluate the improvement effect of RAG on the answer quality of IDAS-LLM, the evaluation metrics used in the fine-tuning comparative experiments are still employed. Focus is primarily on the recall-related metric ROUGE. The comparison of metrics before and after adding RAG is shown in Table 4.

**TABLE 4 Comparison of Metrics Before and After Adding RAG**

| Models | Legal Provision | | | Railway Regulation | | | Railway Expertise | | |
|---|---|---|---|---|---|---|---|---|---|
| | R1 | R2 | RL | R1 | R2 | RL | R1 | R2 | RL |
| Finetuned LLM | 0.47 | **0.25** | 0.40 | 0.39 | 0.20 | 0.32 | 0.38 | 0.16 | 0.28 |
| IDAS-LLM(with RAG) | **0.49** | **0.25** | **0.41** | **0.50** | **0.29** | **0.41** | **0.40** | **0.18** | **0.29** |

From the experimental results, the introduction of RAG significantly improves the framework's performance in answering railway regulation questions, with the three ROUGE metrics each increasing by 11%, 9%, and 9%. This improvement may be related to the cumulative tokens across categories. Since knowledge may overlap among categories, the railway regulations category contains richer information, providing effective prompts for the LLM, thus enabling the generation of higher quality responses. Overall, RAG indeed provides effective references for the LLM's responses, enhancing the quality of the answers.

Additionally, during the indexing phase, railway textual data are processed, chunked, and transformed into embeddings stored in a vector database. The quality of the index built determines whether the correct context can be retrieved during the retrieval phase (*33*). Therefore, documents are split into chunks based on a fixed number of tokens, and evaluation metrics are compared, with results shown in Figure 3.

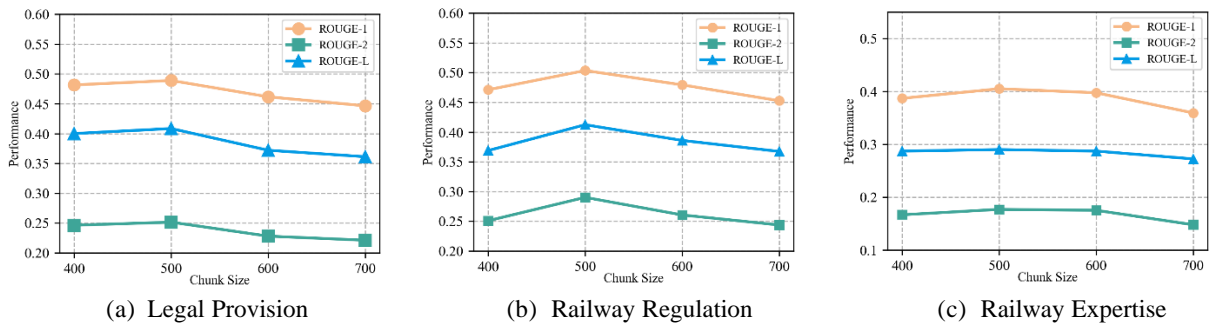

(a) Legal Provision    (b) Railway Regulation    (c) Railway Expertise

**FIGURE 3 Metrics under Different Text Chunk Size**

From the metrics in Figure 3, as they vary with chunk size, it is evident that chunking at 500 tokens yields better metrics across the three categories compared to other chunk sizes. This indicates that in the IDAS-LLM framework, a chunk size of 500 tokens is reasonable for textual data in the driving data knowledge base. Moreover, this also reflects that larger chunks can capture more context but may also introduce more noise. Excessively long text prompts hinder the LLM's ability to summarize and induce, reducing its capacity to extract key information. While smaller chunks may not fully convey the necessary context, they indeed carry less noise. However, chunking can lead to sentence truncation, which is detrimental to generating complete answers.

**CASE STUDY**

To demonstrate the fault handling capabilities of the IDAS-LLM framework during high-speed railway operations, two typical case studies are presented. One involves a common traction loss fault in the train traction drive system, and the other involves a frequently occurring sensor fault in the train braking system.

**Scenario One: Simulation of Traction Loss Faul**

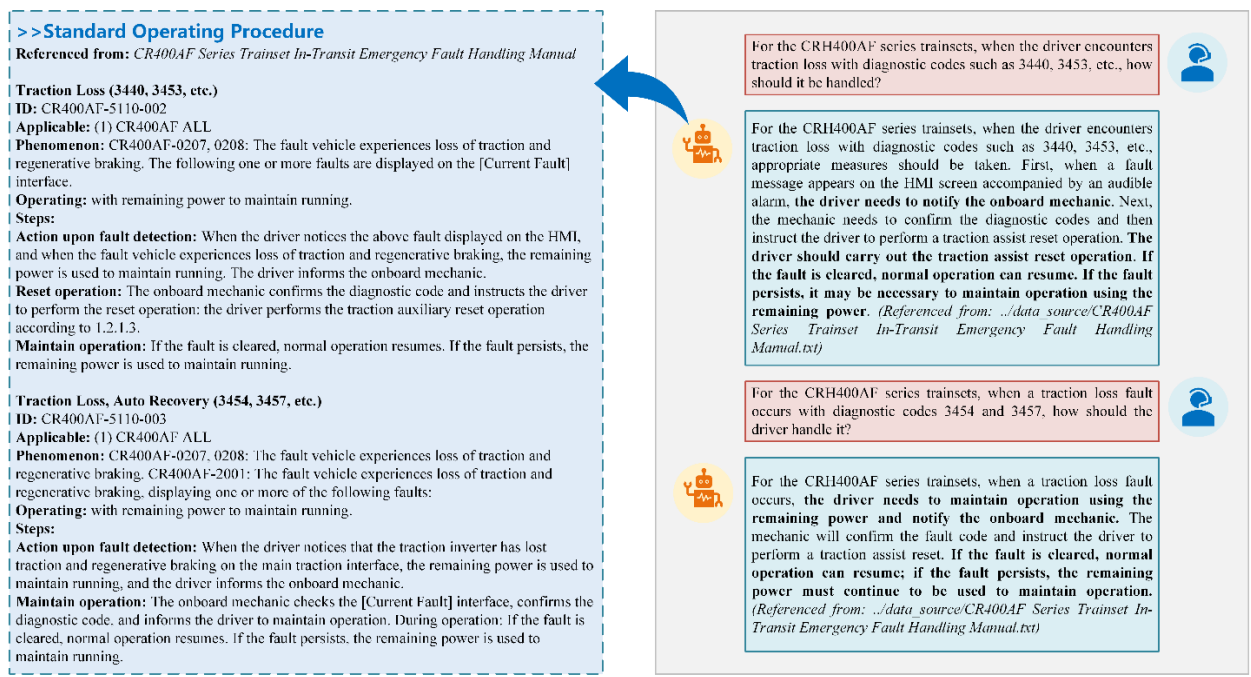

**FIGURE 4 Handling of Different Traction Loss Scenarios Under the Same Train Class**

Traction loss is one of the main faults that occur in the train traction drive system during operation. There are many causes of traction loss, each associated with different diagnostic codes. However, the plethora of diagnostic codes demands a lot from the driver's memory capacity, as drivers must execute appropriate fault handling based on various diagnostic codes. Therefore, two simulation methods are designed. One is for the CR400AF series trainsets, where different diagnostic codes and descriptions are given to demonstrate the fault handling capabilities of IDAS-LLM for various traction loss issues, as shown in Figure 4. The other compares the fault handling recommendations of IDAS-LLM for traction loss across different train classes, as shown in Figure 5, with the CR400AF series and CRH3C / CRH380B (L) series trainsets selected for comparison.

From the responses in Figure 4, it is clear that IDAS-LLM thoroughly references the content of retrieved documents similar in nature, summarizing and stating the information to provide satisfactory handling suggestions for three different traction loss scenarios under the CR400AF series trainsets.

**FIGURE 5 Handling of Traction Loss Scenarios Across Different Train Classes**

From the responses in Figure 5, it is apparent that IDAS-LLM retrieves the emergency fault handling manuals corresponding to the different train classes, CR400AF and CRH3C / CRH380B(L). Based on the retrieved content, IDAS-LLM summarizes and provides effective suggestions.

In general, for traction loss fault scenarios, IDAS-LLM is capable of retrieving the relevant emergency fault handling manuals based on the provided diagnostic codes or phenomena observed in the driver's cab. By combining the retrieved content with the user's actual needs, it provides fault handling suggestions that have practical reference value.

**Scenario Two: Simulation of Sensor Fault**

Sensor faults are also among the most common issues in various systems during train operations. Given the high operating speeds of trainsets, it is crucial to promptly inform the driver and the on-board mechanic in the event of a fault. Consequently, numerous sensors are installed on trains to immediately report any malfunctions detected. Similarly, two simulation methods are designed. One is for the CR400AF series trainsets, where different types of sensor faults and corresponding diagnostic codes are provided to demonstrate IDAS-LLM's retrieval and handling capabilities in the face of various sensor faults, as shown in Figure 6. The other method compares the fault handling suggestions of IDAS-LLM for speed sensors across different train classes, as shown in Figure 7, with the CR400AF series and CRH380B(L) series trainsets selected for comparison.

From the responses in Figure 6, it is evident that under the CR400AF model, IDAS-LLM

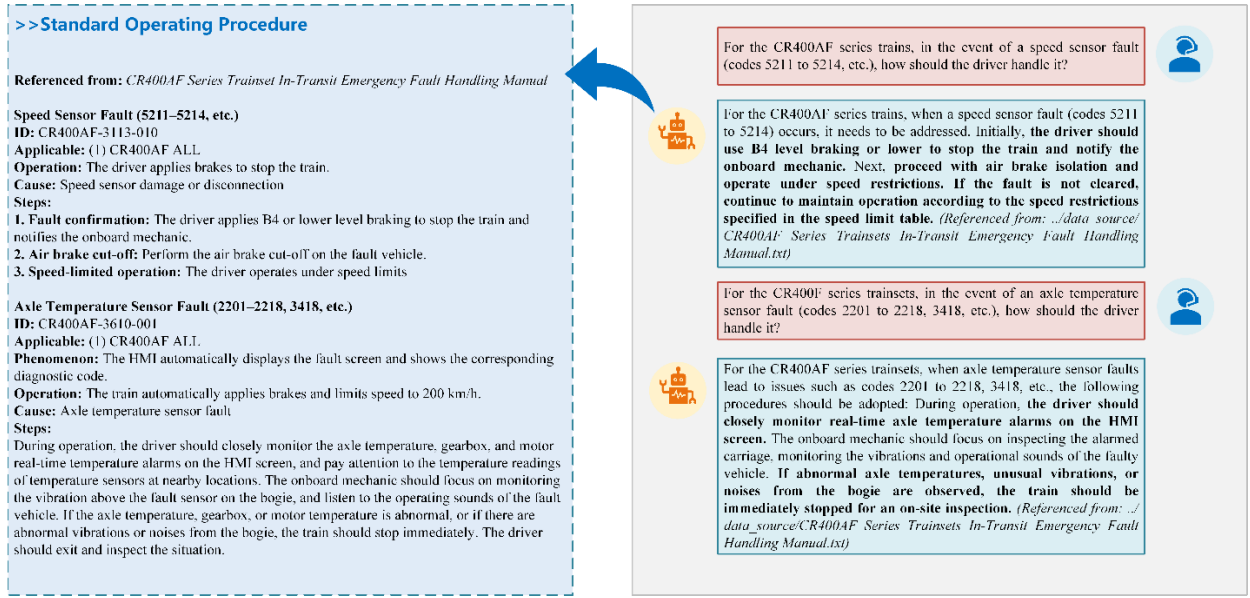

**FIGURE 6 Handling of Different Sensor Faults Under the Same Train Class**

accurately retrieves the corresponding emergency fault handling manuals for speed sensor and axle temperature sensor faults, and provides guidance on fault handling based on the content of these manuals.

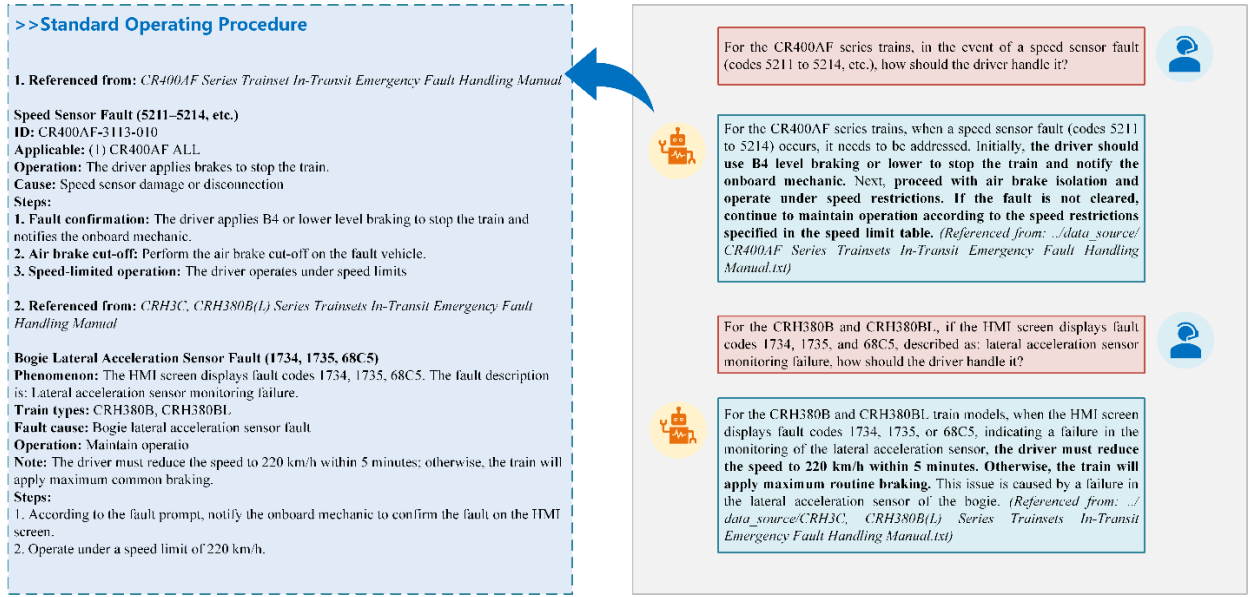

**FIGURE 7 Handling of Speed Sensor Faults Across Different Train Classes**

From the responses in Figure 7, it is manifest that IDAS-LLM retrieves the handling methods for speed sensor faults under two different train classes, assisting drivers in completing the overall fault handling process.

Overall, the IDAS-LLM system demonstrates specific adaptability in handling sensor fault scenarios. It can effectively retrieve the appropriate emergency fault handling manuals based on the specific train class and type of fault sensor. Additionally, the system integrates the search results with the user's specific needs to provide targeted fault handling.

## CONCLUSION

In this work, the IDAS-LLM framework, a large language model based system designed to assist in managing frequent high-speed train faults, is proposed. By incorporating fine-tuning and RAG, the LLM's knowledge base in the railway domain and the explainability of its responses are enhanced, equipping it with fault handling capabilities in real operational scenarios and providing a valuable tool for train drivers. Despite these achievements, the framework's limitations still exist. Although fine-tuning and RAG alleviate LLM hallucinations, they cannot guarantee the accuracy of all responses, and driving suggestions provided in text form may lead to issues such as distracting train drivers.

In the future, efforts can be directed towards creating more realistic and effective railway datasets and incorporating work related to knowledge graphs to further enhance the system's accuracy and explainability. Additionally, exploring how to maintain the system's real-time performance while improving accuracy is a worthwhile direction to investigate.

## ACKNOWLEDGEMENTS

This work is supported by National Key R&D Program of China 2024YFE0104400, Beijing Natural Science Foundation L231028.

## AUTHOR CONTRIBUTIONS

The authors confirm contribution to the paper as follows: study conception and design: J. Xun, Y. C. Luo; data collection: Y. C. Luo, R. Z. Zhang, W. Wang; analysis and interpretation of results: Y. C. Luo, J. Xun; draft manuscript preparation: Y. C. Luo, J. Xun, Z. C. Zhao. All authors reviewed the results and approved the final version of the manuscript.